\documentclass[10pt,twocolumn,letterpaper]{article}
\usepackage{array}
\usepackage{multirow}
\usepackage{makecell}
\usepackage{colortbl}
\usepackage{comment}
\usepackage[pagenumbers]{cvpr} 

\usepackage{lineno}
\definecolor{cvprblue}{rgb}{0.21,0.49,0.74}
\definecolor{mygray}{gray}{0.6}
\usepackage[pagebackref,breaklinks,colorlinks,allcolors=cvprblue]{hyperref}

\def\paperID{4465} 
\def\confName{CVPR}
\def\confYear{2025}

\title{Inst3D-LMM: Instance-Aware 3D Scene Understanding \\ with Multi-modal Instruction Tuning}

\author{ 
  Hanxun Yu$^1$\footnotemark[1], \ \ Wentong Li$^2$\thanks{Equal contribution.}, \ \ Song Wang$^1$, \ \ Junbo Chen$^3$\footnotemark[2], \ \ Jianke Zhu$^1$\thanks{Corresponding authors.} \\[0.1cm]
  {
        $^1$Zhejiang University \ \ \
        $^2$Nanjing University of Aeronautics and Astronautics \ \ \
        $^3$Udeer.ai} \\  
    {\tt\small \{hanxun.yu, songw, jkzhu\}@zju.edu.cn, wentong\_li@nuaa.edu.cn, junbo@udeer.ai}
}

\begin{document}
\maketitle

\begin{figure*}[h]
\centering
\includegraphics[width=1.0\linewidth]{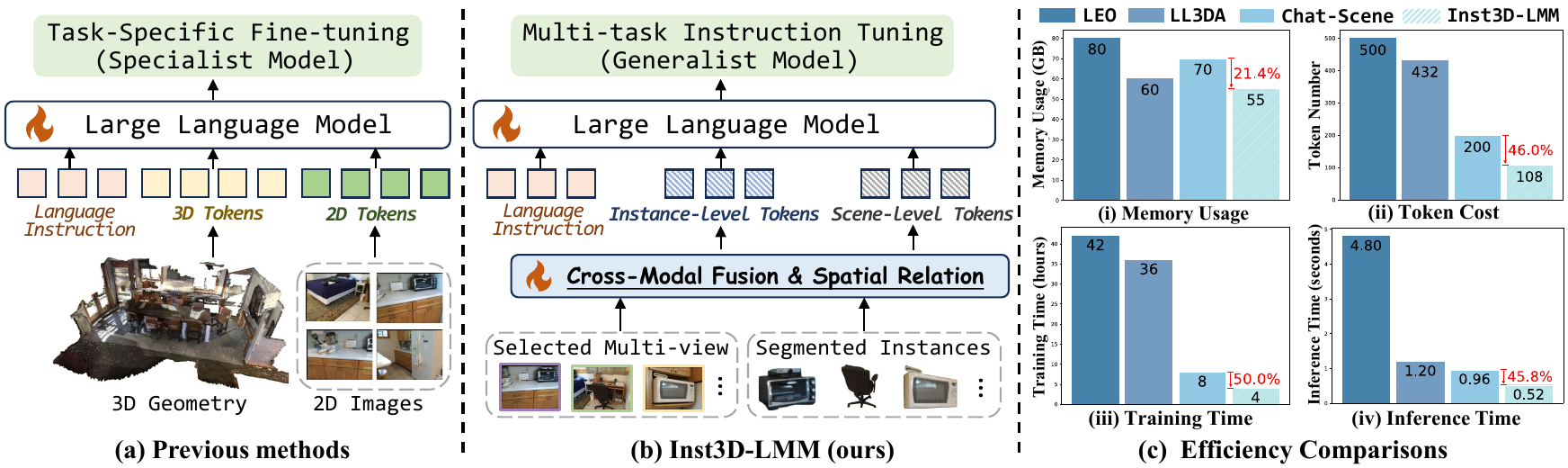} 
\vspace{-5mm}
\caption{\textbf{Comparisons between previous 3D LMMs and our proposed Inst3D-LMM.} (a) Previous methods~\cite{hong20233d,huang2023leo,chen2024ll3da,huang2024chat} typically encode the features of 3D points or 2D images separately and concatenate them directly, often requiring task-specific fine-tuning for different tasks. (b) Illustration of Inst3D-LMM. Our method integrates 2D/3D cross-modal information and captures the intricate spatial relations among objects within 3D environments to generate tight but informative instance/scene-level tokens for the LMM. (c) Compared with other 3D LMMs, our Inst3D-LMM requires fewer computational resources, while offering faster training and inference speeds.}
\label{teaser}
\vspace{-3mm}
\end{figure*}

\begin{abstract}
Despite encouraging progress in 3D scene understanding, it remains challenging to develop an effective Large Multi-modal Model (LMM) that is capable of understanding and reasoning in complex 3D environments. 
Most previous methods typically encode 3D point and 2D image features separately, neglecting interactions between 2D semantics and 3D object properties, as well as the spatial relationships within the 3D environment. This limitation not only hinders comprehensive representations of 3D scene, but also compromises training and inference efficiency.
To address these challenges, we propose a unified \textbf{Inst}ance-aware \textbf{3D} \textbf{L}arge \textbf{M}ulti-modal \textbf{M}odel (Inst3D-LMM) to deal with multiple 3D scene understanding tasks simultaneously. 
To obtain the fine-grained instance-level visual tokens, we first introduce a novel Multi-view Cross-Modal Fusion (MCMF) module to inject the multi-view 2D semantics into their corresponding 3D geometric features. For scene-level relation-aware tokens, we further present a 3D Instance Spatial Relation (3D-ISR) module to capture the intricate pairwise spatial relationships among objects. Additionally, we perform end-to-end multi-task instruction tuning simultaneously without the subsequent task-specific fine-tuning. 
Extensive experiments demonstrate that our approach outperforms the state-of-the-art methods across 3D scene understanding, reasoning and grounding tasks. Source code is available at: \href{https://github.com/hanxunyu/Inst3D-LMM}{https://github.com/hanxunyu/Inst3D-LMM}
 .
\end{abstract}    
\section{Introduction}
\label{sec:intro}

Building Large Multi-modal Models (LMMs) for 3D scene understanding becomes an emerging research topic with significant potential for advancing autonomous robotics~\cite{rana2023sayplan}. For example, the interactive embodied agents~\cite{song2023llm} are expected to interpret 3D layouts and predict object locations based on human instructions.

Traditional 3D scene understanding methods \cite{ma2022sqa3d,wu2023eda} are typically tailored for individual downstream tasks, such as 3D Visual Grounding (3D-VG), 3D Question Answering (3D-QA) and 3D Dense Captioning (3D-DC). 
In contrast, LMMs are able to handle various 3D perception tasks within a single model. 
Some methods~\cite{hong20233d,wang2024embodiedscan} primarily focus on translating 3D points into the space of 2D Vision Language Models (VLMs) or directly leveraging multi-view 2D features as 3D representations. Alternatively, other approaches~\cite{wang2023chat,zhu2024empowering,chen2024ll3da} directly encode the features of 3D points and facilitate the alignment with LLM using 3D-text instruction data.
However, they often require multi-stage alignment or language-scene pre-training, complicating the development of a versatile model capable of handling multiple tasks.
To enable a unified 3D LMM framework, recent work~\cite{huang2024chat} decomposes the input 3D scene into a set of individual object proposals, each identified by unique tokens to capture instance-level 3D object features explicitly.
While this approach exhibits promising results, it neglects the interactions between the 2D semantic features and the properties of 3D objects, as well as the spatial relationship modeling among objects in 3D environments. This oversight further results in substantial token costs for the LLM, thereby hindering both training and inference efficiency.



In this paper, we propose Inst3D-LMM, an effective \textbf{Inst}ance-aware \textbf{3D} \textbf{L}arge \textbf{M}ulti-modal \textbf{M}odel that tackles multiple 3D-language tasks without resorting to task-specific fine-tuning.
Our approach fully leverages the powerful 2D Vision Foundation Models (VFMs) and 3D specialist models to extract enriched 2D and 3D features at the instance level respectively. 
As shown in Figure~\ref{teaser}, in contrast to previous methods, our approach is able to generate fine-grained instance-level representations that encapsulate both geometric and semantic properties,  and scene-level representations that capture intricate pairwise spatial relationships among objects in a 3D scene. Moreover, our method results in minor token costs for the LLM, thereby enhancing both training and inference efficiency.  By leveraging this instance-aware methodology, our Inst3D-LMM significantly improves the LLM's ability to comprehend 3D scenes with respect to both efficiency and accuracy.

Specifically, we introduce a novel Multi-view Cross-Modal Fusion (MCMF) module that effectively infuses enriched multi-view 2D features into the original 3D features with coarse semantics. A learnable \texttt{[CLS]} token is introduced to aggregate the characteristics of each 2D view, enabling efficient multi-view 2D-to-3D cross-modal transformation. To capture intricate pairwise spatial relationships among objects in a 3D scene, we then propose a 3D Instance Spatial Relation (3D-ISR) module.  A spatial condition self-attention between manifold position embeddings and instance-level tokens is presented to produce relation-aware scene-level representations. The resulting instance-level and scene-level representations are subsequently fed into the LLM for end-to-end multi-task instruction tuning.

Extensive experiments across various tasks, including 3D-VG, 3D-QA and 3D-DC, demonstrate that our approach outperforms previous state-of-the-art methods with leading 3D scene understanding, grounding and reasoning capabilities. 
Unlike most existing methods that focus on close-set scene understanding or require per-task fine-tuning, our Inst3D-LMM operates as a generalist model. 
We believe this work lays a fundamental step towards unifying diverse 3D vision-language tasks in generative language modeling.

To summarize, our contributions are as follows:
\begin{itemize}
    \item We propose a unified and efficient instance-aware LLM-based framework, called Inst3D-LMM for various 3D scene understanding tasks with end-to-end multi-modal instruction tuning. Serving as a generalist model, our approach demonstrates superior performance across 3D scene understanding, reasoning and spatial localization.
    \item We utilize 2D VFMs to extract mutli-view contextual features for each 3D instance and then devise a Multi-view Cross-Modal Fusion (MCMF) module to effectively enhance instance-level feature representations by jointly integrating 3D geometry and 2D semantic priors.
    \item 
    A 3D Instance Spatial Relation (3D-ISR) module is introduced to boost the capability of LMM in understanding the complex spatial details within 3D scenes. 
\end{itemize}

\section{Related Works}
\label{sec:related_works}

\begin{figure*}[t]
\centering
\includegraphics[width=1.0\linewidth]{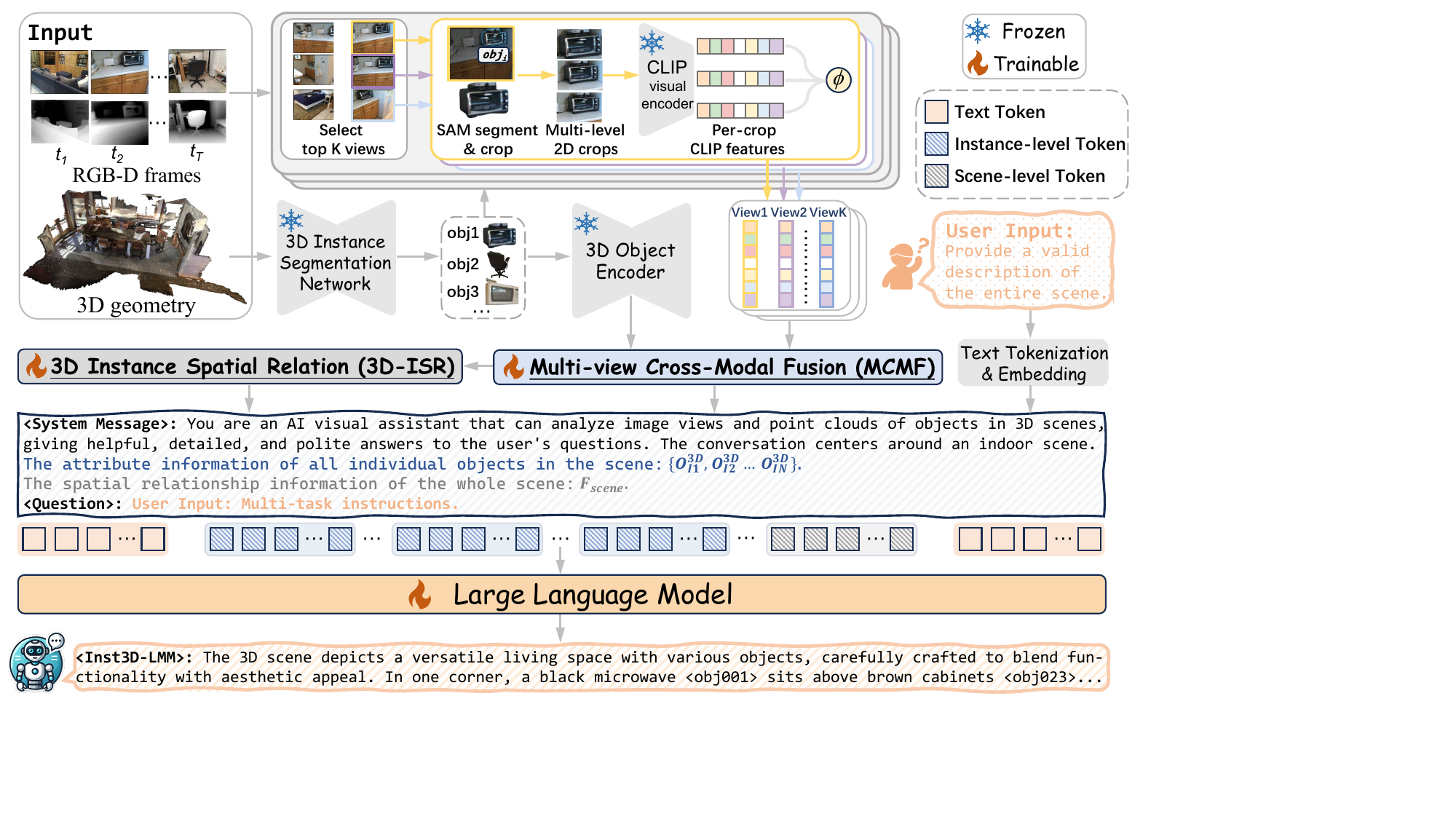} 
\vspace{-5mm}
\caption{\textbf{Overview of our proposed Inst3D-LMM.} Our pipeline takes as input point clouds of a 3D indoor scene, along with RGB-D images. We first employ the pre-trained 3D specialist models and 2D VFMs to extract 3D proposals and multi-view 2D semantic features, respectively. We then suggest the MCMF module to generate fine-grained instance-level tokens. A 3D-ISR module is further introduced to create relation-aware scene-level tokens based on spatial distances. By leveraging the constructed 3D-language  prompts, we conduct multi-task instruction tuning to simultaneously handle various 3D tasks.}
\label{pipeline}
\vspace{-3mm}
\end{figure*}

\noindent \textbf{3D Scene Understanding with Language.}
In 3D scene understanding, there is a surge of interest in making use of language queries to capture user intentions for various downstream tasks, such as 3D Visual Grounding~\cite{chen2022language,wang20233drp}, 3D Question Answering \cite{ma2022sqa3d,parelli2023clip} and 3D Dense Captioning \cite{chen2021scan2cap,jiao2022more}.
Specifically, 3D Visual Grounding entails localizing target objects based on language queries. Moreover, 3D Question Answering demands robust 3D spatial perception and reasoning. 3D Dense Captioning involves localizing and describing objects in 3D scenes. 
The conventional methods typically focus on a specific task. Instead, 3D visual grounding and dense captioning tasks are combined by leveraging their complementary aspects \cite{cai20223djcg,chen2023unit3d,man2024lexicon3d,xu2024vlm}. Recent efforts like 3D-VLP \cite{jin2023context} and 3D-VisTA \cite{zhu20233d} attempt to establish a universal framework by pre-aligning 3D scenes with their corresponding textual descriptions. In contrast to our Inst3D-LMM, most existing methods still focus on closed-set scene understanding, which requires either task-specific fine-tuning or striving to build specialized models.

\noindent \textbf{3D Large Multi-modal Models.}
Inspired by the significant advancements in Large Language Models (LLMs), researchers extend LLM's knowledge to encompass 3D modality \cite{zhou2023uni3d,liu2024openshape,zhu2024llava,yang2024llmi3d}. Point-LLM \cite{guo2023point} and Imagebind-LLM \cite{han2023imagebind} have succeeded in bridging the gap between 3D visuals and text by utilizing extensive 3D object datasets. However, these models struggle with interpreting complex spatial relationships in 3D scenes. Another promising direction focuses on developing the scene-level 3D LMMs. Hong et al.~\cite{hong20233d} encodes projected 3D features using a 2D vision encoder and incorporates location tokens to augment LLM vocabularies. Other methods \cite{wang2023chat,chen2024ll3da,zhu2024empowering} directly encode point clouds and utilize 3D scene-text data for better visual interaction through pre-alignment. Huang et al. \cite{huang2023chat} employs a three-stage training scheme and adopts object identifiers to learn individual object attributes. Chen et al. \cite{chen2024grounded} leverages special referent tokens for precise referencing and grounding. Wang et al. \cite{wang2024embodiedscan} encodes point clouds and RGB images separately. However, some crucial semantic pixels are often lost during their sparse fusion processes~\cite{wei2023moire,wei2024physical}, resulting in a coarse visual semantic representation. In this work, we propose an effective instance-aware framework to fuse fine-grained cross-modal information and encode spatial relations, which achieves promising results on multiple 3D-language tasks as a generalist model. 
\section{Methodology}

Our goal is to enable LLM to understand the 3D environment and perform various visual interaction tasks based on human instructions. Figure~\ref{pipeline} illustrates the architecture of our framework. In this section, we first introduce how to extract features at the instance level using pre-trained 3D models~\cite{schult2023mask3d,zhou2023uni3d} and 2D VFMs~\cite{kirillov2023segment,radford2021learning}, respectively. Secondly, we present a novel Multi-view Cross-Modal Fusion (\textbf{MCMF}) module to obtain fine-grained instance-level tokens, which is specially designed to effectively integrate 3D geometric features with their corresponding multi-view 2D semantic features. Thirdly, we introduce the 3D Instance Spatial Relation (\textbf{3D-ISR}) module to enhance LLM's ability to capture spatial information at the scene level, which generates relation-aware tokens through attention-based analysis of spatial relationships among different 3D proposals. Under our proposed framework, the MCMF and 3D-ISR modules are jointly optimized, enabling mutual enhancement. Finally, we conduct end-to-end multi-task instruction tuning to address a range of 3D scene understanding tasks simultaneously.

\subsection{Instance-Level Feature Extraction}
\noindent \textbf{3D Feature Extraction.} We first segment the 3D point clouds of each individual instance in a \textit{class-agnostic} manner by leveraging a pre-trained 3D instance segmentation model~\cite{schult2023mask3d}. 
We only retain the predicted binary 3D instance masks while ignoring their closed-vocabulary class labels. All instance proposals in one scene are represented by $O^{3D}=\left( O_{1}^{3D}, \ldots, O_{N}^{3D} \right)$, where $O_{i}^{3D}=[coordinate,color]$ consists of the attributes of each instance. We obtain all instance proposals of a scene, and the pre-trained 3D encoder $\mathbf{E_{o}}$~\cite{zhou2023uni3d} is used to extract their instance-level features, \textit{i.e.}, $\mathbf{f_{o}^{3D}} = \mathbf{E_{o}}\left(O^{3D}\right)$.
\\
\noindent \textbf{2D Feature Extraction.} Due to the inherent sparsity of 3D point cloud data, the previous methods have difficulties in generating discriminative features of each object. In this work, we employ powerful 2D VFMs to extract 2D semantic features for each 3D instance. 
As in~\cite{takmaz2023openmask3d,cao2024coda}, we firstly project the point cloud of each instance $O_{i}^{3D}$ onto the image plane and then select the top $K$ views according to the number of visible points. To obtain the accurate 2D masks, we randomly sample $\mathit{k}_{\text{sample}}=5$ points as the input prompts for SAM~\cite{kirillov2023segment} to deal with noisy bounding boxes with outliers. Hereby, we select the high-quality mask with the highest confidence score. To enrich features with contextual information, these masks are employed to generate multi-level ($\mathit{L}$) crops of the selected images that are further fed into the pre-trained CLIP vision encoder ~\cite{radford2021learning} to extract features with language-aligned embedding space. Finally, we aggregate multi-level features of sub-images from the same frame to form the 2D multi-view features $O^{2D}$.

\subsection{Multi-view Cross-Modal Fusion}
\begin{figure}[t]
        \centering
        \includegraphics[width=1.0\linewidth]{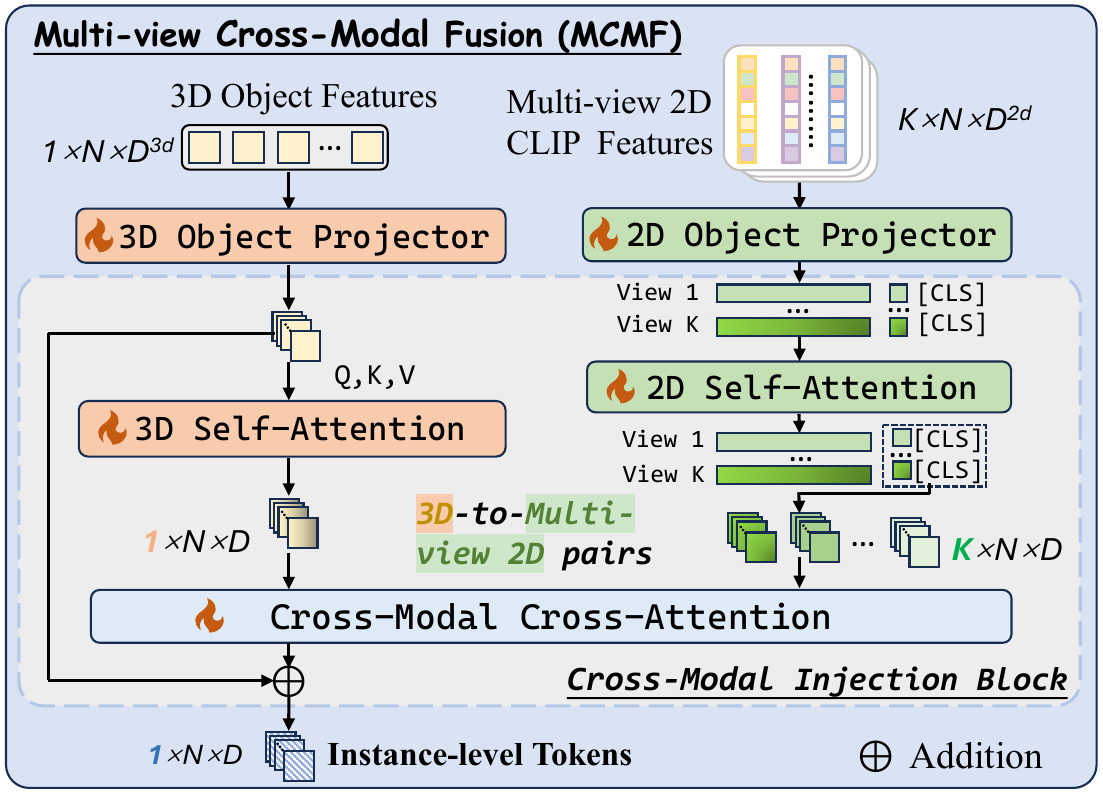}
        \caption{Architecture of the proposed Multi-view Cross-Modal Fusion (MCMF) module.}
        \label{MCMF_module}
        \vspace{-4.8mm}
\end{figure}

To better fuse 3D geometry priors with 2D multi-view semantic priors, we introduce an effective Multi-view Cross-Modal Fusion (MCMF) module that generates the enriched token representations for each 3D instance before being fed into LLM. The architecture of MCMF is crafted with a coarse-to-fine framework, as shown in Figure~\ref{MCMF_module}.
To map 3D object features $O^{3D}\in \mathbb{R}^{1 \times N \times {D^{3d}}}$ and 2D CLIP features $O^{2D} \in \mathbb{R}^{K \times N \times {D^{2d}}}$ into the embedding space of LLM with the dimensionality of $D$, we utilize a simple two-layer MLP with a LayerNorm and GELU in between, yielding  $O^{3D'}$ and $O^{2D'}$, respectively. $N$ represents the number of instances, and $K$ is the number of images for each object.

Subsequently, we introduce a Cross-Modal Injection Block to transform the enriched semantic priors from 2D multi-view representations into 3D instance features. 
Moreover, we adopt a straightforward self-attention layer to further enhance such 3D instance features. For 2D multi-view features, we append a learnable \texttt{[CLS]} token $t_k$ to the flattened CLIP feature maps in order to adaptively encapsulate the global semantic representation of the $k$-th view. Then, we apply a self-attention layer to multi-view features, enabling semantic gathering to derive $t_{k}^{'}$ for $k$-th view as:
\begin{equation}
    t_{k}^{'} = \phi(\text{SelfAttn}([t_k, O_{\text{view}\_k}^{2D'}])),
\end{equation}
where $1 \leq k \leq K$, 
$O_{\text{view}\_k}^{2D'} \in \mathbb{R}^{ N \times {D}}$. $\phi$ is an indicator function that outputs the updated \texttt{[CLS]} token as the first one from the token list.

We obtain the processed 3D token embedding $O^{3D'}\in \mathbb{R}^{1 \times N \times D}$ containing coarse semantics, where each instance in $O^{3D'}$ corresponds to multiple views of the processed 2D CLIP features $O^{2D'}$. It can be represented as follows:
\begin{equation}
   O_i^{2D'} = \left[ t_1', \ldots, t_K' \right], \quad i \in [1, N].
\end{equation}
To infuse the enriched semantic 2D features into 3D geometric feature, we construct the 3D-to-2D-Multi-view pairs, \textit{i.e.} each 3D instance feature in $O^{3D'}\in \mathbb{R}^{1 \times N \times D}$ corresponds to 2D features across $K$ views in $O^{2D'}\in \mathbb{R}^{K \times N \times D}$. In particular, we utilize the preliminary $O^{3D'}$ as queries, while the 2D multi-view visual features $O^{2D'}$ as enriched reference keys and values. Inspired by~\cite{li2024tokenpacker}, 
the injection process is conducted via a cross-attention layer, which encourages  3D instance queries to absorb the fine-grained semantics of keys and values. This results in the enhanced 3D instance features $O^{3D}_{f} \in \mathbb{R}^{1 \times N \times D}$ as below:
\begin{equation}
    O_{f}^{3D} = \text{CrossAttn}(O^{3D'},O^{2D'}).
\end{equation}
Furthermore, we leverage a residual operation by adding the input $O^{3D'}$ to retain the basic characteristics in generating the instance-level 3D visual tokens, $O_{I}^{3D} = O_{f}^{3D} + O^{3D'}$.

\begin{figure}[t]
        \centering
        \includegraphics[width=1.0\linewidth]{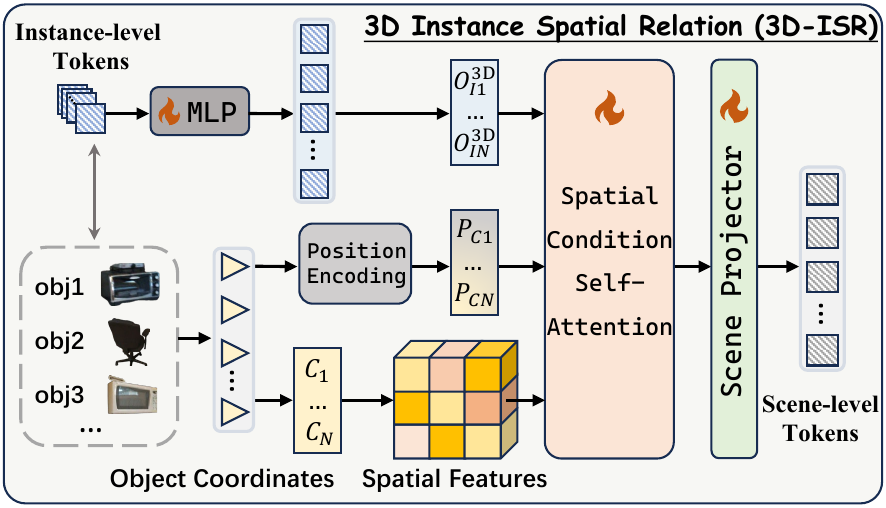}
        \caption{Illustration of the 3D Instance Spatial Relation (3D-ISR) module in our framework.}
        \label{3D-ISR}
        \vspace{-4mm}
\end{figure}

\subsection{3D Instance Spatial Relation}
Motivated by~\cite{chen2022language,zhu20233d,liangspatioawaregrouding3d}, we develop an effective 3D Instance Spatial Relation (3D-ISR) module to boost LLM's capabilities in assimilating spatial information within the 3D scene, as shown in Figure~\ref{3D-ISR}. 3D-ISR utilizes the instance-level 3D visual tokens $\{O_{I1}^{3D},O_{I2}^{3D}, \ldots, O_{IN}^{3D}\}$ derived from MCMF module along with the corresponding center coordinates $\{C_{1},C_{2}, \ldots, C_{N}\}$ of all instances as inputs. For the $i$-th instance, we define its center coordinates as $C_i = \left(x_i,y_i,z_i\right)$.
For each pair of instance-level tokens $\{O_{Ii}^{3D},O_{Ij}^{3D}\}$, we calculate their Euclidean distance $d_{ij}=\left\| C_i - C_j \right\|_2$ as well as their horizontal angle $\theta_h$ and vertical angle $\theta_v$. Specially,  $\theta_h = \arctan2({{({y_j} - {y_i})} \mathord{\left/
 {\vphantom {{({y_j} - {y_i})} {({x_j} - {x_i})}}} \right.
 \kern-\nulldelimiterspace} {({x_j} - {x_i})}})$ and $\theta_v = \arcsin (({{{z_j} - {z_i})} \mathord{\left/
 {\vphantom {{{z_j} - {z_i})} {{d_{ij}}}}} \right.
 \kern-\nulldelimiterspace} {{d_{ij}}}})$. By making use of these positional parameters, we generate the pairwise spatial features $S=\{s_{ij}\} \in \mathbb{R}^{N \times N \times 5}$ as:
\begin{gather}
    s_{ij} = \left[\sin(\theta_h), \cos(\theta_h), \sin(\theta_v), \cos(\theta_v), d_{ij}\right].
\label{eq:spatial_features}
\end{gather}

Inspired by language-conditioned self-attention presented in  ViL3DRel~\cite{chen2022language}, we suggest a spatial-conditioned self-attention module. Specially, we create position embeddings $\mathcal{P}$ via the absolution positional encoding (PE), \textit{i.e.}  $\mathcal{P} = \text{PE}(C_{1},C_{2},\ldots,C_{N})$. 
A spatial conditioned attention weight $l_i$ is computed to select the relevant spatial relations for each instance $O_{Ii}^{3D}$, which is formulated as below:
\begin{gather}
    l_i = W_P^\top (\mathcal{P}_{i} + O_{Ii}^{3D}),
\end{gather}
where $W_P \in \mathbb{R}^{D \times 5}$ is a learnable parameter and the bias term is omitted for simplicity. The spatial-conditioned attention map $\omega_{ij}$ is computed by combining $s_{ij}$, $l_i$ and $l_j$, \textit{i.e.} $\omega_{ij} = l_{i} \cdot s_{ij} \cdot l_{j}$.

Through this positional transformation, the spatial-conditioned attention map $\omega_{ij}$ encapsulates the pairwise spatial relationship across the entire 3D scene for each instance. We further integrate the attention map with instance-level visual tokens, which is formulated as below:
\begin{gather}
    F_i = \sum_{j=1}^{N} \omega_{ij} O_{Ii}^{3D}.
\end{gather}
Finally, the instance-level representations are concatenated and further processed by a transformer-based encoder $\Gamma$, followed by a max-pooling layer (Pool) and a simple two-layer MLP. The scene-level relation-aware tokens $F_{scene} \in \mathbb{R}^{D}$ are generated 
 for the entire 3D scene as follows:
\begin{gather}
    F_{scene} = \text{MLP}(\text{Pool}(\Gamma([F_1, F_2, \ldots, F_N]))).
\end{gather}

Thus,  we obtain the final token representations for LLM processing, \textit{i.e.} instance-level visual tokens $O_{Ii}^{3D}$ and spatial relation-aware scene-level tokens $F_{scene}$, respectively.

\subsection{End-to-End Multi-task Instruction Tuning}
\noindent \textbf{3D-Language Prompts.} Given an LMM, clear and explicit system messages and instructions are essential to support a range of downstream 3D vision-language tasks. 
For different tasks, we adopt various task-specific instruction templates with 3D features to generate uniform instruction data, enabling multi-task training. 
\textit{Please refer to the Supplementary Material for more detailed information.}

\begin{figure*}[t]
\centering
\includegraphics[width=0.98\linewidth]{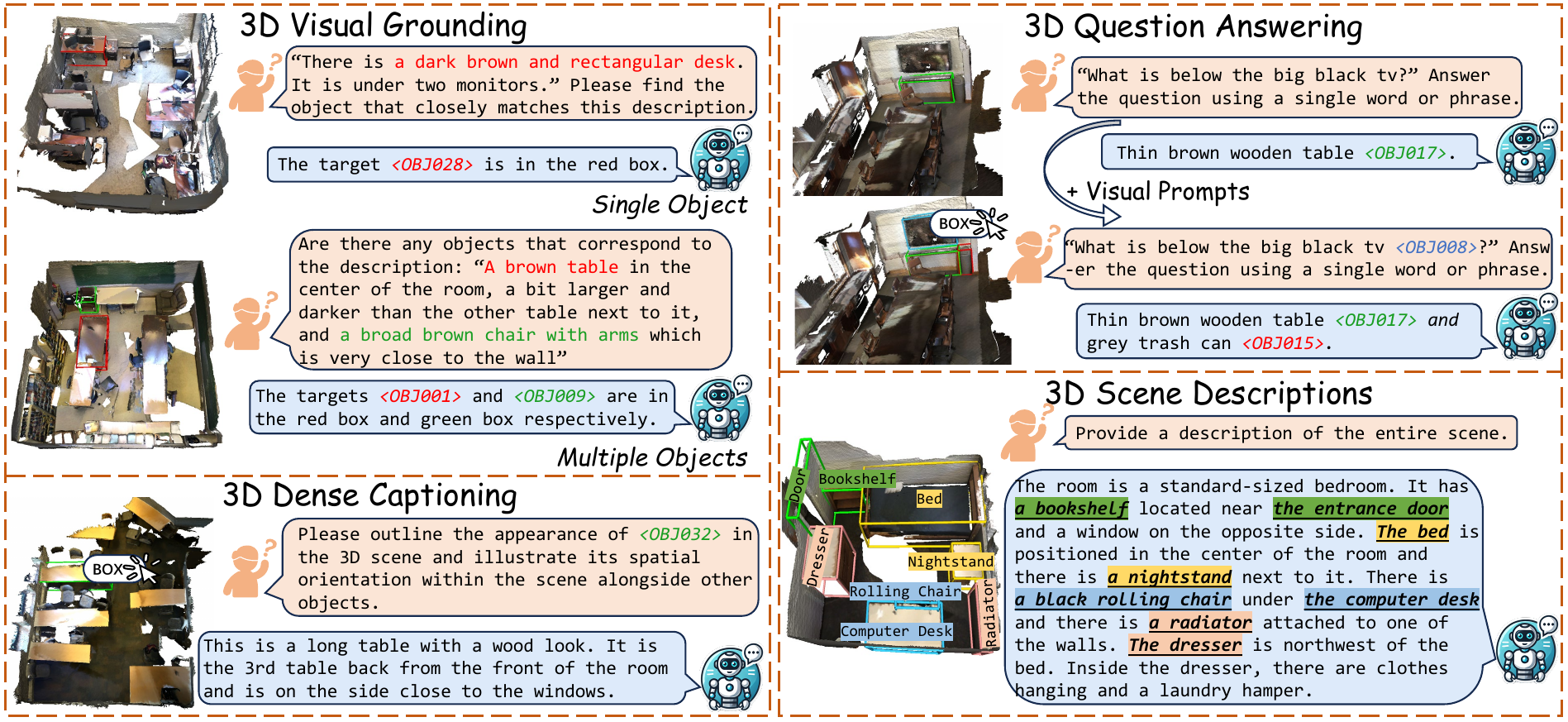} 
\vspace{-1mm}
\caption{Qualitative illustration of Inst3D-LMM across various 3D-language tasks in diverse 3D environments.} 
\vspace{-4mm}
\label{all_tasks}
\end{figure*}

\noindent \textbf{Instruction Tuning.} 
We conduct end-to-end multi-task fine-tuning to fully leverage the capabilities of our Inst3D-LMM framework. The MCMF and 3D-ISR modules enable the LLM with robust 3D scene understanding, grounding and reasoning abilities. We freeze the pre-trained 3D object encoder while updating both MCMF and 3D-ISR modules, as well as LLM.  
Upon completion of end-to-end multi-task instruction tuning, Inst3D-LMM can effectively handle various 3D vision-language tasks simultaneously, without fine-tuning on specific tasks, as illustrated in Figure~\ref{all_tasks}.


\section{Experiments}
\label{sec:experiments}


\definecolor{mygreen}{RGB}{0, 128, 0}
\definecolor{myblue}{RGB}{108, 160, 220}
\definecolor{mygray}{gray}{0.6}
\begin{table*}[t]
    \setlength{\tabcolsep}{4.4pt}
    \centering
    \renewcommand\arraystretch{0.88}
    \scalebox{0.88}{%
    \begin{tabular}{lcccccc|ccc}
        \toprule
        \multirow{2}{*}{Method} & \multicolumn{2}{c}{Overall} & \multicolumn{2}{c}{Unique} & \multicolumn{2}{c|}{Multiple} & \multicolumn{2}{c}{Multi3DRefer} \\
        \cmidrule(lr){2-3} \cmidrule(lr){4-5} \cmidrule(lr){6-7} \cmidrule(lr){8-9}
         & Acc@0.25\textuparrow & Acc@0.50\textuparrow & Acc@0.25\textuparrow & Acc@0.50\textuparrow & Acc@0.25\textuparrow & Acc@0.50\textuparrow & F1@0.25\textuparrow & F1@0.50\textuparrow \\
        \midrule
        \midrule
        \textbf{\textit{Closed-set, full sup.}} \\
        \textcolor{mygray}{ScanRefer \cite{chen2020scanrefer}} &  \textcolor{mygray}{37.3}& \textcolor{mygray}{24.3}& \textcolor{mygray}{65.0}&\textcolor{mygray}{43.3}& \textcolor{mygray}{30.6}&\textcolor{mygray}{19.8}&\textcolor{mygray}{--}&\textcolor{mygray}{--}\\
        
        \textcolor{mygray}{3DVG-Trans \cite{zhao20213dvg}} & \textcolor{mygray}{45.9}& \textcolor{mygray}{34.5}& \textcolor{mygray}{77.2}&\textcolor{mygray}{58.5}& \textcolor{mygray}{38.4}&\textcolor{mygray}{28.7}&\textcolor{mygray}{30.2}&\textcolor{mygray}{25.5}\\


        

        \textcolor{mygray}{M3DRef-CLIP \cite{zhang2023multi3drefer}} &  \textcolor{mygray}{51.9}& \textcolor{mygray}{44.7}& \textcolor{mygray}{--}&\textcolor{mygray}{77.2}& \textcolor{mygray}{--}&\textcolor{mygray}{36.8}&\textcolor{mygray}{42.8}&\textcolor{mygray}{38.4}\\
        
        \textcolor{mygray}{ConcreteNet \cite{unal2023three}} &  \textcolor{mygray}{56.1}& \textcolor{mygray}{49.5}& \textcolor{mygray}{86.1}&\textcolor{mygray}{79.2}& \textcolor{mygray}{47.5}&\textcolor{mygray}{40.9}&\textcolor{mygray}{--}&\textcolor{mygray}{--}\\
        

        \textcolor{mygray}{CORE-3DVG \cite{yang2024exploiting}} & \textcolor{mygray}{56.8}& \textcolor{mygray}{43.8}& \textcolor{mygray}{85.0}&\textcolor{mygray}{67.1}& \textcolor{mygray}{51.8}&\textcolor{mygray}{39.8}&\textcolor{mygray}{--}&\textcolor{mygray}{--}\\
        \midrule
        \textbf{\textit{Zero-shot}} \\
        LLM-Grounder \cite{yang2023llm} & 17.1& 5.3& 30.8& 22.6&16.3&12.9&--&--\\
        Visual-Programming \cite{yuan2024visual} & 36.4& 32.7& \underline{63.8} & \underline{58.4} & \underline{27.7} & \underline{24.6}&--&--\\
        \midrule
        \textbf{\textit{Specialist}}\\
        OpenScene \cite{peng2023openscene} & 13.0& 5.1& 20.1& 13.1& 11.1& 4.4&--&--\\
        3D-LLM(Flamingo) \cite{hong20233d} & 21.2& --& --& --&--&--&--&--\\
        3D-LLM(BLIP2-flant5) \cite{hong20233d} & 30.3& --& --& --&--&--&--&--\\
        Chat-3D v2 \cite{huang2023chat} & 35.9& 30.4& 61.2& 57.6&25.2&22.6&45.1&41.6\\
         ReGround3D \cite{zhu2024empowering} & 53.1 & 41.1 & --& --&--&--&--&--\\
        \midrule
        \textbf{\textit{Generalist}}\\
        LAMM \cite{yin2024lamm} & --& 3.4& --& --&--&--&--&--\\
        3DMIT \cite{li20243dmit} & 10.7& 7.2& --& --&--&--&--&--\\
        Grounded 3D-LLM \cite{chen2024grounded} & 47.9 & 44.1 & --& --&--&--&45.2&40.6\\
        Chat-Scene \cite{huang2024chat} & \underline{55.5}& \underline{50.2}& -- & -- & -- & -- & \underline{57.1}&\underline{52.4}\\
        \rowcolor{gray!30}
        Inst3D-LMM  & \textbf{57.8} & \textbf{51.6} & \textbf{88.6} & \textbf{81.5}& \textbf{48.7} &\textbf{43.2}&\textbf{58.3}&\textbf{53.5}\\
        \bottomrule
    \end{tabular}
    }
     \vspace{-2mm}
    \caption{Quantitative results for 3D Visual Grounding on ScanRefer and Multi3DRefer validation sets. In the ScanRefer dataset, scenes are labeled as ``unique" (one object per class) or ``multiple" (more than one).
    Closed-set methods are fully supervised for specific datasets.
    ``Zero-shot" refers to methods that directly use LLMs without fine-tuning. ``Specialist" and ``Generalist" categorize methods fine-tuned for specific tasks versus those trained jointly. \textbf{Bold} and \underline{underlined} numbers indicate the best and the second-best results, respectively.}
    \label{tab:scanrefer}
    \vspace{-1mm}
\end{table*}

\begin{table*}[t]
    \centering
    \scalebox{0.785}{%
    \begin{tabular}{lccccccccccc}
        \toprule
        \multirow{2}{*}{Method} & \multirow{2}{*}{\makecell{\# 3D Data \\ for Alignment}}  & \multicolumn{6}{c}{ScanQA} & \multicolumn{4}{c}{Scan2Cap@0.50}\\
        \cmidrule(lr){3-8} \cmidrule(lr){9-12}
        &   & B-1\textuparrow & B-4\textuparrow & METEOR\textuparrow & ROUGE\textuparrow &  CIDER\textuparrow & EM\textuparrow & B-4\textuparrow & METEOR\textuparrow & ROUGE\textuparrow & CIDER\textuparrow  \\
        
        \midrule
        \midrule
        \textbf{\textit{Closed-set, full sup.}}\\
        \textcolor{mygray}{VoteNet \cite{ding2019votenet} + MCAN \cite{yu2019deep}} &  \textcolor{mygray}{--} &  \textcolor{mygray}{28.0}& \textcolor{mygray}{6.2}& \textcolor{mygray}{11.4}&\textcolor{mygray}{29.8}& \textcolor{mygray}{54.7}&\textcolor{mygray}{17.3}&\textcolor{mygray}{--}&\textcolor{mygray}{--}&\textcolor{mygray}{--}&\textcolor{mygray}{--}\\

        \textcolor{mygray}{ScanRefer \cite{chen2020scanrefer} + MCAN \cite{yu2019deep}} & \textcolor{mygray}{--} &  \textcolor{mygray}{26.9}& \textcolor{mygray}{7.9}& \textcolor{mygray}{11.5}&\textcolor{mygray}{30.0}& \textcolor{mygray}{55.4}&\textcolor{mygray}{18.6}&\textcolor{mygray}{--}&\textcolor{mygray}{--}&\textcolor{mygray}{--}&\textcolor{mygray}{--}\\

        \textcolor{mygray}{ScanQA \cite{chen2021scan2cap}} & \textcolor{mygray}{--} &  \textcolor{mygray}{30.2}& \textcolor{mygray}{10.1}& \textcolor{mygray}{13.1}&\textcolor{mygray}{33.3}& \textcolor{mygray}{64.9}&\textcolor{mygray}{21.0}&\textcolor{mygray}{--}&\textcolor{mygray}{--}&\textcolor{mygray}{--}&\textcolor{mygray}{--}\\
        
        \textcolor{mygray}{Scan2Cap \cite{azuma2022scanqa}}  & \textcolor{mygray}{--} &  \textcolor{mygray}{--}& \textcolor{mygray}{--}& \textcolor{mygray}{--}&\textcolor{mygray}{--}& \textcolor{mygray}{--}&\textcolor{mygray}{--}&\textcolor{mygray}{22.4}&\textcolor{mygray}{21.4}&\textcolor{mygray}{43.5}&\textcolor{mygray}{35.2}\\
        
        \textcolor{mygray}{3D-VisTA \cite{zhu20233d}}  & \textcolor{mygray}{--} &  \textcolor{mygray}{34.2}& \textcolor{mygray}{13.1}& \textcolor{mygray}{15.2}&\textcolor{mygray}{38.6}& \textcolor{mygray}{76.6}&\textcolor{mygray}{27.0}&\textcolor{mygray}{34.0}&\textcolor{mygray}{27.1}&\textcolor{mygray}{54.3}&\textcolor{mygray}{66.9}\\        
        \midrule
        \textbf{\textit{LLM-based Methods}}\\
        LLaVA (zero-shot) \cite{liu2024visual} &  -- &  7.1 & 0.3& 10.5 & 12.3 & 5.7 & 0.2 & 1.5 & 8.3 & 19.6&3.2  \\
        LAMM \cite{yin2024lamm}  & 25K & 26.8& 5.8& 10.0& 23.6&42.4&9.8&--&--&--&--\\
        3D-LLM(Flamingo) \cite{hong20233d}  & 675K & 30.3& 7.2& 12.2& 32.3&59.2&20.4&5.9&11.4&29.9&--\\
        3D-LLM(BLIP2-flant5) \cite{hong20233d}  & 675K & \underline{39.3} & 12.0& 14.5& 35.7&69.4&20.5&8.1&13.1&33.2&--\\
        Chat-3D v2 \cite{huang2023chat}  & 38K & 38.4& 7.3& \underline{16.1}& \underline{40.1}&77.1&\underline{21.1}& 31.8 & 22.3 & 50.2&63.9\\
        LL3DA \cite{chen2024ll3da}  & 38K & --& 13.3& 15.4 & 37.0 & 75.7&--& 35.9 & \underline{25.6} & \underline{54.6}&65.2 \\
        Grounded 3D-LLM \cite{chen2024grounded}  & 107K & --& 13.4 & --& --&72.7&--&  35.5 & -- & -- & 70.6 \\
        Chat-Scene \cite{huang2024chat}  & 38K & -- & \underline{14.3} & -- & --
 &\underline{87.7}& -- & \underline{36.3} & -- & -- &\underline{77.1}\\
        \rowcolor{gray!30}
        Inst3D-LMM  & 38K & \textbf{43.5} & \textbf{14.9} & \textbf{18.4} & \textbf{42.6} & \textbf{88.6} & \textbf{24.6} &  \textbf{38.3} & \textbf{27.5} & \textbf{57.2}&\textbf{79.7} \\
        \bottomrule
    \end{tabular}
    }
    \vspace{-1mm}
    \caption{Quantitative results for 3D Question Answering and 3D Dense Captioning on the ScanQA and Scan2Cap datasets.}
    \label{tab:scanqa}
    \vspace{-3.0mm}
\end{table*}

\subsection{Experimental Setting}
\noindent \textbf{Datasets and Benchmarks.} In this work, we conduct our experiments on the ScanNetv2 dataset \cite{dai2017scannet}, an extensive indoor 3D scene dataset comprising 1,513 scenes. This dataset includes 3D point clouds, RGB-D frames, and detailed point-level instance segmentation annotations. 
The whole dataset is divided into 1,201 scenes for training and 312 scenes for validation, with all subsequent benchmarks adhering to these training/validation splits. 
Our evaluation encompasses a range of 3D scene understanding benchmarks, including ScanRefer \cite{chen2020scanrefer} and Multi3DRefer \cite{zhang2023multi3drefer} for single- and multi-object 3D Visual Grounding, respectively, ScanQA \cite{azuma2022scanqa} for 3D Question Answering, and Scan2Cap \cite{chen2021scan2cap} for 3D Dense Captioning. These datasets are converted into a uniform instruction format for multi-task instruction tuning and performance assessment.

\noindent \textbf{Implementation Details.} For 3D feature extraction, we leverage the 3D instance segmentation model Mask3D~\cite{schult2023mask3d} pre-trained on ScanNet200 dataset~\cite{rozenberszki2022language}, alongside the 3D object encoder Ulip2~\cite{xue2023ulip2}/Uni3D~\cite{zhou2023uni3d} based on ViT-L/14~\cite{dosovitskiy2020image}. 
For 2D feature extraction, we adopt the ViT-H-based SAM~\cite{kirillov2023segment} to obtain high-quality masks. Moreover, we exact  2D semantic features using the vision encoder from CLIP-ViT-L/14-336px~\cite{radford2021learning}. These pre-trained models are kept frozen.
In our method, we apply multi-view selection and multi-level crops to 2D images, setting $K=5$ views and $L=3$ levels. We use Vicuna1.5-7B \cite{chiang2023vicuna} as our basic LLM, which is fine-tuned from LLaMA2 \cite{touvron2023llama}. 
Our fine-tuning process utilizes LoRA \cite{hu2021lora}. We adopt the AdamW optimizer with a weight decay of 0.02. All experiments are conducted on 8 NVIDIA A100 GPUs.

\subsection{Main Results}
\begin{figure}[t]
        \centering
        \includegraphics[width=1.0\linewidth]{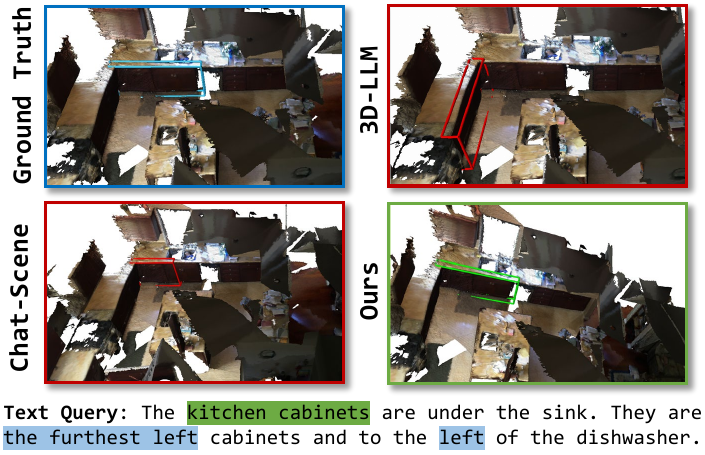}
        \vspace{-4mm}
        \caption{Visual comparisons in 3D Visual Grounding. The rendered images show the ground-truth (\textcolor{blue}{blue}), incorrectly identified objects (\textcolor{red}{red}), and correctly identified objects (\textcolor{green}{green}). The colored text indicates the results of text decoupling.}
        \label{small-vg}
        \vspace{-5.0mm}
\end{figure}

\noindent \textbf{3D Visual Grounding.} 
We first report the visual grounding performance on ScanRefer and Multi3DRefer validation datasets. As shown in Table~\ref{tab:scanrefer}, our approach outperforms the state-of-the-art model, Chat-Scene~\cite{huang2024chat}, by \textcolor{mygreen}{+2.3}\% Acc@0.25 and \textcolor{mygreen}{+1.2}\% F1@0.25 on ScanRefer and Multi3DRefer, respectively. Compared to Specialist and closed-set methods, Inst3D-LMM, \textit{trained with a generalist approach}, achieves competitive performance. Figure~\ref{small-vg} displays typical visual comparison results.\\
\noindent \textbf{3D Question Answering.} We then compare  Inst3D-LMM with previous leading methods on the ScanQA validation set. Table~\ref{tab:scanqa} reports the results. Our Inst3D-LMM consistently outperforms these methods, including the recent LL3DA~\cite{chen2024ll3da}, Grounded 3D-LLM~\cite{chen2024grounded} and Chat-Scene~\cite{huang2024chat}.\\
\noindent \textbf{3D Dense Captioning.} This task involves the localization and description of instances. As shown in Table~\ref{tab:scanqa}, our approach achieves 38.3\% B-4@0.50 and 79.7\% C@0.50 on Scan2Cap, which  exceeds the closed-set expert model 3D-VisTA~\cite{zhu20233d} by \textcolor{mygreen}{+4.3}\% and \textcolor{mygreen}{+12.8}\%, and still outperforms LLM-based method LL3DA by \textcolor{mygreen}{+2.4}\% and \textcolor{mygreen}{+14.5}\%.

\subsection{Ablation Study}
In this section, we perform ablation experiments to thoroughly evaluate the effectiveness of components. \textit{Please refer to the Supplementary Material for more analysis.} \\
\textbf{Effects of Multi-view Cross-Modal Fusion (MCMF).} The MCMF module aims to augment LLM's understanding of 3D instances' geometric and semantic attributes. We explore several methods to combine these features, including direct concatenation for projection (`Concat.'), parallel projection followed by combination (`Parallel'), and vanilla cross-attention (`Cross-Attention'). To encode the basic spatial information, we concatenate the results of absolution positional encoding (PE) with the 3D features. As shown in Table~\ref{tab:ablation_MCMF},  MCMF outperforms all these methods on ScanRefer, ScanQA and Scan2Cap datasets. We also observe that the performance of 3D LMM is significantly improved by incorporating multi-view 2D CLIP information compared to using only 3D geometric features. These results demonstrate the effectiveness of our MCMF approach.


\begin{table}[h]
\vspace{-2mm}
\centering
\resizebox{0.48\textwidth}{!}{%
\begin{tabular}{c|cccccccccc}
\toprule
\multirow{2}{*}{Method} & \multicolumn{2}{c}{ScanRefer} & \multicolumn{2}{c}{ScanQA} & \multicolumn{2}{c}{Scan2Cap@0.50} \\
\cmidrule(lr){2-3} \cmidrule(lr){4-5} \cmidrule(lr){6-7}
                    &Acc@0.25\textuparrow & Acc@0.50\textuparrow & B-1\textuparrow & CIDER\textuparrow & B-4\textuparrow & CIDER\textuparrow \\ 
                    \midrule \midrule
\textit{w/o} Multi-view 2D & 36.0 & 31.9 & 31.5 & 57.3 & 19.9 & 54.3  \\ 
 Concat. & 38.8 & 34.4 & 36.6 & 65.4 & 25.2 & 63.3\\ 
 Parallel & 37.6 & 33.5 & 35.3 & 63.9 & 24.8 & 62.1  \\ 
 Cross-Attention & 39.2 & 36.5 & 37.4 & 66.1 & 25.5 & 64.3  \\
 \rowcolor{gray!30}
 MCMF & \textbf{46.7} & \textbf{41.9} & \textbf{41.5} & \textbf{78.6} & \textbf{32.7} & \textbf{68.2} \\
\bottomrule
\end{tabular}
}
\vspace{-1.5mm}
\caption{Ablation evaluations of the proposed Multi-view Cross-Modal Fusion (MCMF) module.}
\vspace{-4mm}
\label{tab:ablation_MCMF}
\end{table}

\begin{figure}[t]
        \centering
        \includegraphics[width=1.0\linewidth]{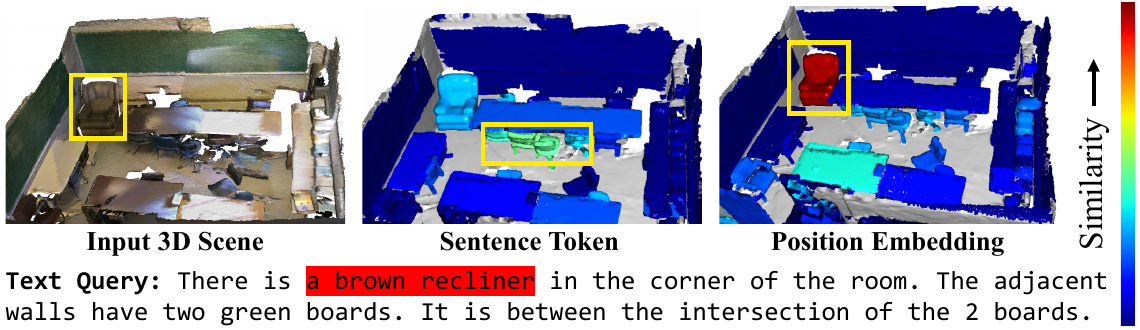}
        \vspace{-4mm}
        \caption{A visualization of the similarity score between text query and segmented 3D proposals. We compare sentence tokens used in ViL3DRel~\cite{chen2022language} with position embeddings employed in 3D-ISR.}
        \label{3d-isr-similarity}
        \vspace{-5mm}
\end{figure}

\noindent \textbf{Impacts of 3D Instance Spatial Relation (3D-ISR).} We further investigate 3D-ISR module to analyze its impact on 3D spatial understanding using only 3D geometry priors.
We compare 3D-ISR against previous spatial relation modeling methods, such as the 3D localization mechanism in ViL3DRel~\cite{chen2022language} and 3D-LLM~\cite{hong20233d}, and relation-aware token generation in Chat-3D-v2~\cite{huang2023chat}. Table~\ref{tab:ablation_3D-ISR} reports the comparison results. It can be observed that our 3D-ISR consistently surpasses these approaches.
Specifically, 3D-ISR significantly enhances performance on grounding tasks (ScanRefer and Multi3dRefer). This aligns with the motivation of our design.  We also provide a visualization, as shown in Figure~\ref{3d-isr-similarity}, of the similarity between the text query and segmented 3D proposals after processing by LLM. Compared to ViL3DRel~\cite{chen2022language} using sentence tokens, our 3D-ISR more accurately captures spatial relationships within the scene by employing position embeddings of objects.
\begin{table}[h]
\vspace{-2.0mm}
\centering
\resizebox{0.48\textwidth}{!}{%
\begin{tabular}{c|cccccc}
\toprule
\multirow{2}{*}{Method} & \multicolumn{2}{c}{ScanRefer} & \multicolumn{2}{c}{Multi3DRefer} & \multicolumn{2}{c}{ScanQA} \\
\cmidrule(lr){2-3} \cmidrule(lr){4-5} \cmidrule(lr){6-7}
                     & Acc@0.25\textuparrow    & Acc@0.50\textuparrow & F1@0.25\textuparrow  & F1@0.50\textuparrow    & B-1\textuparrow & CIDER\textuparrow   \\ \midrule \midrule
\textit{w/o} spatial relation & 35.6 & 29.8 & 34.8 & 27.3 & 29.2 & 55.6 \\
 ViL3DRel \cite{chen2022language}& 36.1 & 30.5 & 38.4 & 35.5 & 32.8 & 58.9 \\
 3D-LLM \cite{hong20233d} & 39.2 & 36.8 & 42.4 & 35.2 & 32.5 & 60.2 \\ 
 Chat-3D-v2 \cite{huang2023chat}& 40.8 & 37.5 & 41.6 & 37.8 & 35.1 & 64.6  \\ 
 \rowcolor{gray!30}
 3D-ISR  & \textbf{48.3} & \textbf{44.1} & \textbf{46.2} & \textbf{41.3} & \textbf{39.1} & \textbf{72.3}  \\
\bottomrule
\end{tabular}
}
\vspace{-2.5mm}
\caption{Ablation study to verify the effectiveness of our 3D Instance Spatial Relation (3D-ISR) module.}
\vspace{-3mm}
\label{tab:ablation_3D-ISR}
\end{table}

\noindent \textbf{Ablations within MCMF and 3D-ISR modules.} We also explore the effectiveness of key designs in MCMF and 3D-ISR modules. The learnable \texttt{[CLS]} token in the MCMF module is introduced to aggregate 2D multi-view features. We compare it against alternative methods, such as token max pooling and Q-Former~\cite{li2023blip}.  Table~\ref{tab:ablation_cls_token} reports the comparison results, demonstrating the efficacy of the learnable \texttt{[CLS]} token design. Besides, to assess the impact of different pairwise spatial features in the 3D-ISR module, we evaluate the model using only distance or orientation information to compute spatial features in Eq~\ref{eq:spatial_features}. Results in Table~\ref{tab:ablation_in_3D-ISR} reveal that distance has a greater impact on the model's grounding capability, while orientation is more crucial in handling Q\&A tasks. Combining both pairwise distance and orientation yields the best overall performance.

\noindent \textbf{Mutual Benefits of MCMF and 3D-ISR.} As shown in Table~\ref{tab:collaboration}, the model integrating both MCMF and 3D-ISR consistently outperforms those utilizing either module alone across multiple tasks. To further verify its effectiveness, we combine MCMF with absolution positional encoding (PE) for the basic spatial relation, and directly adopt feature concatenation (Concat.) to integrate  3D object features and 2D multi-view features along with 3D-ISR. These results indicate the synergy efforts between MCMF and 3D-ISR. 

\begin{table}[t]
\centering
\resizebox{0.48\textwidth}{!}{%
\begin{tabular}{c|cccccccccc}
\toprule
\multirow{2}{*}{Method} & \multicolumn{2}{c}{ScanRefer} & \multicolumn{2}{c}{ScanQA} & \multicolumn{2}{c}{Scan2Cap@0.50} \\
\cmidrule(lr){2-3} \cmidrule(lr){4-5} \cmidrule(lr){6-7}
                    &Acc@0.25\textuparrow & Acc@0.50\textuparrow & B-1\textuparrow & CIDER\textuparrow & B-4\textuparrow & CIDER\textuparrow \\ \midrule \midrule
Max Pooling & 48.7 & 44.6 & 39.2 & 74.8 & 30.6 & 65.5 \\ 
Q-Former~\cite{li2023blip} & 53.2 & 47.9 & \textbf{44.3} & 87.9 & 35.4 & 78.6  \\ 
\rowcolor{gray!30}
CLS Token & \textbf{57.8} & \textbf{51.6} & 43.5 & \textbf{88.6} & \textbf{38.3} & \textbf{79.7} \\
\bottomrule
\end{tabular}
}
\vspace{-2mm}
\caption{Ablation study of learnable CLS token in MCMF module.}
\vspace{-3.5mm}
\label{tab:ablation_cls_token}
\end{table}

\begin{table}[t]
\centering
\resizebox{0.48\textwidth}{!}{%
\begin{tabular}{c|cccccc}
\toprule
\multirow{2}{*}{Method} & \multicolumn{2}{c}{ScanRefer} & \multicolumn{2}{c}{Multi3DRefer} & \multicolumn{2}{c}{ScanQA} \\
\cmidrule(lr){2-3} \cmidrule(lr){4-5} \cmidrule(lr){6-7}
                     & Acc@0.25\textuparrow    & Acc@0.50\textuparrow & F1@0.25\textuparrow  & F1@0.50\textuparrow    & B-1\textuparrow & CIDER\textuparrow   \\ \midrule \midrule
\textit{w/} Dist (only) & 43.2 & 38.0 & 42.4 & 37.8 & 31.2 & 65.7 \\
 \textit{w/} Ort (only) & 39.7 & 33.6 & 40.5 & 35.9 & 32.5 & 66.8 \\
 \rowcolor{gray!30}
 Dist + Ort & \textbf{48.3} & \textbf{44.1} & \textbf{46.2} & \textbf{41.3} & \textbf{39.1} & \textbf{72.3} \\ 
\bottomrule
\end{tabular}
}
\vspace{-1.5mm}
\caption{Ablation study of 3D-ISR module. Dist (Ort) means only using distance (orientation) to compute pairwise spatial features.}
\vspace{-5.5mm}
\label{tab:ablation_in_3D-ISR}
\end{table}

\noindent \textbf{Different LLMs and Foundation Models.} Table~\ref{tab:ablation_LLM_mini} presents  the results of various LLMs and pre-trained 2D/3D models, including Vicuna-7B~\cite{chiang2023vicuna} \textit{vs} Vicuna-13B, Ulip2~\cite{xue2023ulip2} \textit{vs} Uni3D~\cite{zhou2023uni3d}, CLIP~\cite{radford2021learning} \textit{vs} SigLIP~\cite{mu2022slip}, and ViT-H-based SAM~\cite{kirillov2023segment} \textit{vs} ViT-L-based SAM. We find that the performance of grounding and reasoning increases along with the total number of parameters in the foundation models (\textit{i.e.} ViT-L \textit{vs} Vit-H and 7B \textit{vs} 13B). These results indicate that our framework's capabilities can be improved in tandem with the performance of foundation models. 

\begin{table}[t]
\centering
\resizebox{0.472\textwidth}{!}{%
\begin{tabular}{c|cccccc}
\toprule
\multirow{2}{*}{Method} & \multicolumn{2}{c}{ScanRefer} & \multicolumn{2}{c}{ScanQA} & \multicolumn{2}{c}{Scan2Cap@0.50} \\
\cmidrule(lr){2-3} \cmidrule(lr){4-5} \cmidrule(lr){6-7}
                     & Acc@0.25\textuparrow    & Acc@0.50\textuparrow & B-1\textuparrow & CIDER\textuparrow & B-4\textuparrow & CIDER\textuparrow  \\ \midrule\midrule
\textit{w/} MCMF (only)& 38.2 & 34.6 & 39.4 & 76.8 & 30.5 & 66.9\\
\textit{w/} 3D-ISR (only) & 48.3 & 44.1 & 39.1 & 72.3 & 29.8 & 64.3\\
MCMF + PE & 46.7 & 41.9 & 41.5 & 78.6 & 32.7 & 68.2 \\
3D-ISR + Concat. & 49.8 & 45.0 & 40.6 & 75.4 & 33.3 & 66.5 \\
\rowcolor{gray!30}
MCMF + 3D-ISR (full) & \textbf{57.8} & \textbf{51.6} & \textbf{43.5} & \textbf{88.6} & \textbf{38.3} & \textbf{79.7}\\
\bottomrule
\end{tabular}
}
\vspace{-1.5mm}
\caption{Comparison results of the collaboration between MCMF and 3D-ISR modules.}
\vspace{-4mm}
\label{tab:collaboration}
\end{table}

\begin{table}[t]
\centering
\resizebox{0.472\textwidth}{!}{
\begin{tabular}{c|c|c|cccc}
\toprule
\multirow{2}{*}{LLM} & \multirow{2}{*}{3D Encoder} & \multirow{2}{*}{2D VFMs} & \multicolumn{2}{c}{ScanRefer} & \multicolumn{2}{c}{ScanQA} \\
\cmidrule(lr){4-5} \cmidrule(lr){6-7} 
 & &  & Acc@0.25\textuparrow & Acc@0.50\textuparrow & B-1\textuparrow & CIDER\textuparrow \\ \midrule\midrule
Vicuna-7B & Ulip2 & CLIP+SAM (ViT-L) & 49.2 & 44.8 & 39.5 & 79.6 \\
Vicuna-7B & Ulip2 & CLIP+SAM (ViT-H) & 53.6 & 46.5 & \textbf{43.7} & 87.5 \\
Vicuna-7B & Ulip2 & SigLIP+SAM (ViT-H) & 54.3 & 47.0 & 42.2 & 84.5\\ \midrule
\rowcolor{gray!30}
Vicuna-7B & Uni3D & CLIP+SAM (ViT-H) & \textbf{57.8} & 51.6 & 43.5 & \textbf{88.6}\\
Vicuna-13B & Uni3D & CLIP+SAM (ViT-H) & 56.0 & \textbf{52.0} & 42.8 & 83.1 \\
Vicuna-13B & Uni3D & SigLIP+SAM (ViT-H) & 55.2 & 47.8& 43.3 & 85.4\\
\bottomrule
\end{tabular}
}
\vspace{-1.5mm}
\caption{Ablation study of different LLMs and pre-trained foundation models.}
\vspace{-4mm}
\label{tab:ablation_LLM_mini}
\end{table}

\begin{table}[h]
\centering
\resizebox{0.472\textwidth}{!}{%
\begin{tabular}{c|c|ccc}
\toprule
{Method} & \#Tokens & {VRAM \textdownarrow} & {Training Time\textdownarrow} & {Inference Time\textdownarrow} \\ 
\midrule\midrule
\textit{w/} Separate Encoding & 6$N$ & $\sim$80 GB  & $\sim$42 hours & $\sim$4.80 seconds\\
\textit{w/} Parallel Projection & 3$N$ & $\sim$70 GB  & $\sim$20 hours & $\sim$2.25 seconds \\
\textit{w/} Cross-Attention & 2$N$ & $\sim$65 GB  & $\sim$8 hours & $\sim$1.58 seconds\\
\textit{w/} MCMF & $N$ & $\sim$58 GB & $\sim$5 hours & $\sim$0.76 seconds\\
\rowcolor{gray!30}
MCMF+3D-ISR & $N$ & $\sim$55 GB & $\sim$4 hours & $\sim$0.52 seconds\\
\bottomrule
\end{tabular}
}
\vspace{-1.5mm}
\caption{More ablation analysis on the contribution of efficiency.}
\vspace{-5.5mm}
\label{tab:efficiency}
\end{table}

\subsection{Efficiency Analysis}
The efficiency improvements of Inst3D-LMM are primarily attributable to the reduction in token counts, given that the computational costs of LLM scale quadratically with the number of input tokens. Unlike previous methods that separately encode and concatenate 2D/3D information, requiring an excessive number of tokens, our MCMF module reduces tokens to one per object proposal, while the 3D-ISR module further minimizes the token requirement for spatial relationships to just eight per 3D scene. This reduction in token count substantially lowers computational costs, thereby accelerating both the training and inference processes. Table~\ref{tab:efficiency} provides empirical evidence that supports the efficacy of our modules, emphasizing the impact of token count reduction on overall efficiency.


\section{Conclusion and Limitations}
\label{sec:conclusion}
In this paper, we proposed Inst3D-LMM, an effective instance-aware framework to leverage the potential of Large Multi-modal Models (LMMs) for 3D scene understanding. To improve instance-level representations, we developed a novel Multi-view Cross-Modal Fusion (MCMF) module, which injects multi-view 2D semantic open-vocabulary priors into 3D geometry features to generate fine-grained instance-level tokens.  Furthermore, we introduced a 3D Instance Spatial Relation (3D-ISR) module that employs the spatial condition attention mechanism to capture pairwise spatial relations. Experimental results demonstrate that our approach achieves promising performance in understanding and reasoning across various 3D vision-language tasks. 
\\
\textbf{Limitations.} Due to the scarcity of high-quality 3D-text datasets, there remains a gap between 3D LMM learning and real-world embodied action control, such as robotic manipulation and navigation. In the future, we plan to enhance Inst3D-LMM's reasoning and planning capabilities by scaling up diverse 3D vision and language data. Additionally, ethical safety concerns and potential hallucinatory outputs in LLM applications also warrant attention. 

{
    \small
    \bibliographystyle{ieeenat_fullname}
    \bibliography{main}
}

\end{document}